\documentclass[journal]{IEEEtran}
\usepackage{cite}

\ifCLASSINFOpdf
\usepackage[pdftex]{graphicx}
\usepackage[caption=false,font=footnotesize,labelfont=sf,textfont=sf]{subfig}
  
\else
\usepackage[caption=false,font=footnotesize]{subfig}
\fi
\usepackage{amsmath}
\usepackage{amsfonts}
\begin{document}
%
\title{ESFNet: Efficient Network for Building Extraction from High-Resolution Aerial Images}
%
%
%

\author{
	\IEEEauthorblockN{
		Jingbo Lin\IEEEauthorrefmark{1},
		Weipeng Jing\IEEEauthorrefmark{2},~\IEEEmembership{Member,~IEEE},
		Houbing Song\IEEEauthorrefmark{3},~\IEEEmembership{Senior,~IEEE}, and
		Guangsheng Chen\IEEEauthorrefmark{4},~
	}

	\IEEEauthorblockA{\IEEEauthorrefmark{1}\IEEEauthorrefmark{2}\IEEEauthorrefmark{4}College of Information and Computer Engineering, Northeast Forestry University, Harbin, HLJ China}
	
	\IEEEauthorblockA{\IEEEauthorrefmark{3}Department of Electrical, Computer, Software, and Systems Engineering, Embry-Riddle Aeronautical University, Daytona Beach, FL 32114 USA}
	\thanks{Manuscript received December 1, 2012; revised August 26, 2015. 
		Corresponding author: M. Shell (email: http://www.michaelshell.org/contact.html).}}

%
%

\markboth{Journal of \LaTeX\ Class Files,~Vol.~14, No.~8, August~2015}%
{Shell \MakeLowercase{\textit{et al.}}: Bare Demo of IEEEtran.cls for IEEE Journals}
%



\maketitle

\begin{abstract}
Building footprint extraction from high-resolution aerial images is always an essential part of urban dynamic monitoring, planning and management. It has also been a challenging task in remote sensing research. In recent years, deep neural networks have made great achievement in improving accuracy of building extraction from remote sensing imagery. However, most of existing approaches usually require large amount of parameters and floating point operations for high accuracy, it leads to high memory consumption and low inference speed which are harmful to research. In this paper, we proposed a novel efficient network named ESFNet which employs separable factorized residual block and utilizes the dilated convolutions, aiming to preserve slight accuracy loss with low computational cost and memory consumption. Our ESFNet obtains a better trade-off between accuracy and efficiency, it can run at over 100 FPS on single Tesla V100, requires 6x fewer FLOPs and has 18x fewer parameters than state-of-the-art real-time architecture ERFNet while preserving similar accuracy without any additional context module, post-processing and pre-trained scheme. We evaluated our networks on WHU Building Dataset and compared it with other state-of-the-art architectures. The result and comprehensive analysis show that our networks are benefit for efficient remote sensing researches, and the idea can be further extended to other areas. The code is public available at: https://github.com/mrluin/ESFNet-Pytorch
\end{abstract}

\begin{IEEEkeywords}
Building Extraction, Deep Learning, Efficient Neural Networks, Remote Sensing, Semantic Segmentation.
\end{IEEEkeywords}

%
\IEEEpeerreviewmaketitle

\section{Introduction}
\begin{figure}[!t]
	\centering
	\subfloat[]{\includegraphics[width=1.1in]{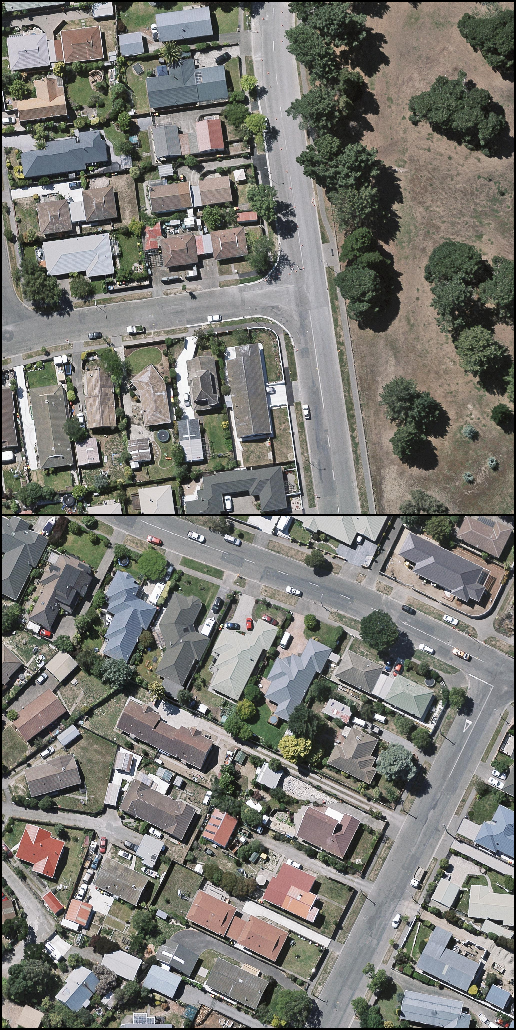}}
	\label{1_a}\hfill
	\subfloat[]{\includegraphics[width=1.1in]{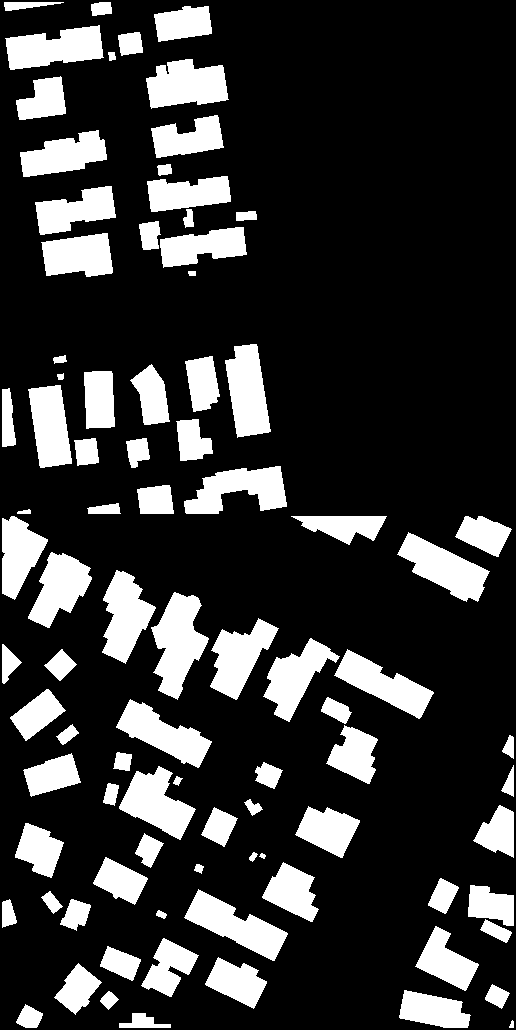}}
	\label{1_b}\hfill 
	\subfloat[]{\includegraphics[width=1.1in]{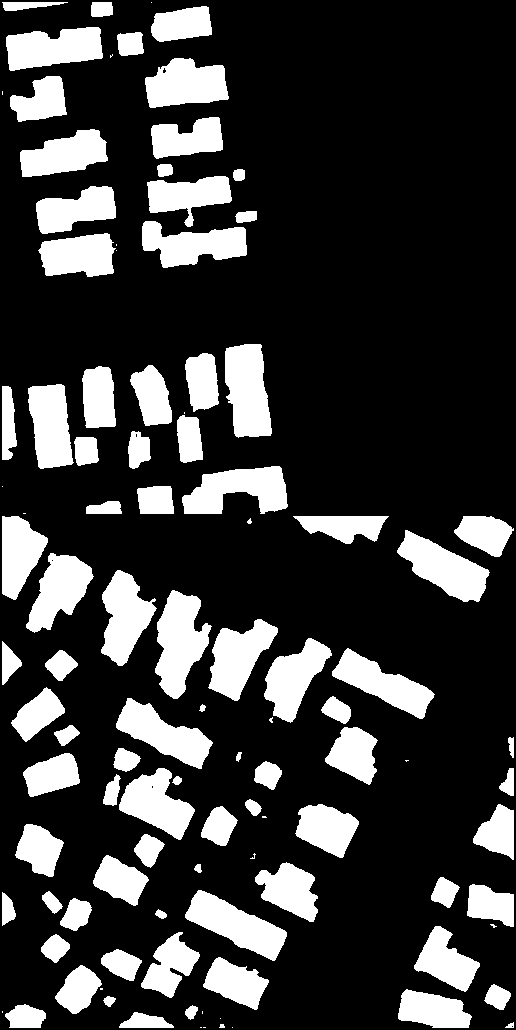}}
	\label{1_c}
	\caption{Examples of building extraction. (a) aerial images. (b) ground truth in which buildings are in white and background is in black. (c) prediction of our network.}
	\label{fig_1}
\end{figure}

%
%
%
%
\IEEEPARstart{H}{igh-resolution} aerial images are widely used in modern smart cities \cite{Song:smart_cities, Zou:smart_forest}. One of the most import applications is automatic building extraction (shown in Fig. \ref{fig_1}) which is aimed to separate pixels belong to buildings with others in urban environments, and it can be considered as pixel-level classification problem, also defined as semantic segmentation task in computer vision. Semantic segmentation of remote sensing imagery has great significance on remote sensing research, such as sea-land segmentation \cite{deepunet}, road detection \cite{automatic road}, and land cover objects classification \cite{scene_classification}.
Nowadays, technology has reached an unprecedented level, the advanced sensors and multimedia systems provide people a better life \cite{jiang:wearable_sensor,jiang:multimedia_iot}. The advanced remote sensing sensors provide more and more high-quality, high-resolution aerial images with much higher ground sampling distance than the past \cite{guided_filter}, so that the imagery usually contains abundant land cover information and confusing environment backgrounds which increased the difficulty on semantic segmentation task especially in urban areas. As a result, traditional learning-based method, which is over dependent upon manual designed features, cannot solve the problems of large scale dataset and meet requirements of nowadays practical applications. 

In the last several years, convolutional neural networks have made great achievement in kinds of computer vision tasks, such as classification \cite{imagenet}, object detection \cite{coco,Yang:aircraft_detection}, image quality retargeting \cite{jiang:image_retargeting} and semantic segmentation \cite{pascal}. Since the milestone work of Long \emph{et al}. \cite{fcn} in 2014, they convert the classical CNN to FCN (fully convolutional neural network) by replacing the fully connected layers into intermediary score maps and using multi-scale feature fusion scheme to solve dense pixel-level classification tasks, the approaches of deep neural networks have been extensively used in semantic segmentation tasks, and they gradually replaced the conventional approaches in which features are extracted manually. Inspired by Ronneberger \emph{et al}. \cite{unet}, encoder-decoder architecture is widely used in segmentation tasks, such as SegNet \cite{segnet}, DeconvNet \cite{deconv} and U-Net \cite{unet}, to name a few. Encoder is usually based on fashion classification networks which is designed to learn high-level semantic representation of the whole imagery, and the decoder is used to match the resolution of output from encoder to the original. For further improving accuracy, DeepLab family \cite{deeplab1, deeplab2, deeplab3, deeplab33} utilize post-processing and additional context module, such as dense conditional random fields (dense-CRF) \cite{deeplab1} and atrous spatial parallel pyramid (ASPP) \cite{deeplab3, deeplab33}. Although these networks significantly improved segmentation performance, they usually require high computational cost, large memory consumption and too much time to train. It is not benefit for scientific research specifically in remote sensing which also needs large memory allocation for high-resolution data with constraint computational resources. So besides accuracy, computation complexity, memory usage and inference speed are also essential metrics to measure the performance of an architecture \cite{shufflenetv2}. Under this intuition, there is a variety of architectures are designed towards high accuracy and efficiency, such as MobileNet family \cite{mobilenet, mobilenetv2}, ShuffleNet family \cite{shufflenet, shufflenetv2}, ENet \cite{enet}, ERFNet \cite{erfnet}, EDANet \cite{edanet} and so on. The trade-off between efficiency and accuracy becomes a key element for designing these efficient architectures.


In order to solve the efficiency limitation which presents in the existing approaches, we proposed a highly improved architecture called ESFNet (Efficient Separable Factorized Network) based on factorized residual block in real-time architecture ENet \cite{enet} and ERFNet \cite{erfnet}. Benefits from depth-wise convolutions, we employed SFRB (Separable Factorized Residual Block) as our core module, it can compress model size and reduce the computation complexity drastically. Considering the characteristics of ResNet \cite{resnet} and the factors that can influence the efficiency of segmentation networks, we extended ESFNet to ESFNet-mini-ex, it performs much better than the base one in inference speed. We evaluated our architecture on recent expressive building dataset WHU Building Dataset, via comparing with the state-of-the-art and comprehensive analysis, elaborating that our model is an efficient backbone for semantic segmentation and the idea can be further extended to other computer vision tasks. In summary, there are three main contributions as follows:

\begin{itemize}
	\item \textbf{} We proposed a novel efficient network named ESFNet,  which employed separable factorized residual block with dilated convolutions. It can run 100.29 FPS on single Tesla V100 and achieve 85.34\% IoU on WHU Building Dataset which is similar to the state-of-the-art. Our ESFNet-mini-ex further increased the inference speed to 142.98 FPS and achieved 84.57\% IoU with only 1\% accuracy loss than the base.
	\item \textbf{} The proposed ESFNet can run 12 more frames per second than novel real-time architecture ERFNet with 6x fewer floating point operations and 18x fewer parameters, and it had only 2\% accuracy loss which can be considered as a proper balance between accuracy and efficiency. Our ESFNet-mini-ex can be better that performed 54 more frames per second with 3\% accuracy loss.
	\item \textbf{} We conducted sets of ablation studies to observe the performance of different architectures and analyzed the reasons behind them.
\end{itemize}

The rest of the paper is organized as follows. In Section II we review related works. Section III introduces each part of our network. Section IV exhibits expressive comparison results and analysis. Section V concludes the whole paper.

\section{Related Works}
There are lots of FCN-based methods used in semantic segmentation and achieved high accuracy, but the existing works usually ignored the efficiency of architecture. As a result most of top-accuracy networks are usually heavy-weight which is harmful for the cases with constraint computational resources. 

Maggiori \emph{et al}. \cite{maggi_fcn_remote} designed a multi-scale neuron module to reduce the trade-off between recognition and localization. But there are lots of large kernels in their network. The large kernel brought large amount of parameters and memory consumption, it is inefficient and can be replaced by stacking small kernels \cite{vgg}. Yuan \cite{yuan_building} tackled dense prediction by integrating activation from multiple layers in different stages. But the VGG-based straight up and down structure is not benefit for information flowing and feature reusing. Li \emph{et al}. \cite{li_reuse} proposed encoder-decoder architecture and employed Dense-block \cite{densenet} as their core module. But the Dense-block makes networks need too much memory access cost. Ji \emph{et al}. \cite{whu_dataset} proposed Siamese U-Net to improve segmentation performance by multi-scale input. But the deep symmetry architecture means it needs heavy-weight decoder which leads to high memory consumption and low inference speed. As a result, although the FCN-based and UNet-based methods make great success on accuracy, they are not suitable for practical applications and efficient remote sensing research.

One way to obtain light-weight networks is utilizing efficient structure, such as residual block, kernel factorization and group convolutions. Recently, there are lots of deep learning approaches managed toward light-weight and real-time \cite{enet,erfnet,edanet,jiang:internet}. ENet provides principles of designing efficient segmentation networks and is one of the first networks designed for light-weight architecture, it employs bottleneck structure and factorized kernels to keep low computational cost and small amount of parameters. An extension of ENet, ERFNet also benefits from residual block and factorized kernels, it gets much better balance between accuracy and efficiency than ENet. EDANet employs asymmetric residual structure, dilated convolutions and the dense connectivity to achieve high efficiency and accuracy. Thus most of recent novel real-time segmentation networks are benefit from residual block whose basic intention is alleviating degradation problem of the deep neural networks, the residual block is also an efficient structure that can speed up the training phase. Group convolution \cite{gc} is an efficient convolution operation widely used in many efficient networks, it divides the input into independent groups and the kernels of each group share the same weight in order to reduce the number of parameters. Other efficient networks benefit from depth-wise convolution \cite{mobilenet,mobilenetv2,shufflenet,shufflenetv2,xception} which is extreme case of group convolution. Recent works such as BiSeNet \cite{bisenet} and ICNet \cite{icnet} also have better trade-off between accuracy and speed, but they are not easy to deploy and difficulty in migrating to other tasks and areas due to their complex structures. Although these networks have already performed well in efficiency, there is still a big room for further improvement.

\begin{table}[!t]
	\renewcommand{\arraystretch}{1.3}
	\caption{The detailed information of our ESFNet. And the input and output are in $H\times C$ format, where we only consider square images and 
		H is spatial size, C is number of channel of the feature maps}
	\label{table_1}
	\centering
	\begin{tabular}{c c c c c}
		\hline \hline 
		\bfseries ID & \bfseries Blocks & \bfseries Input &\bfseries Output \\
		\hline
		1 & Initial Block &                      $512\times 3$  & $256\times 16$  \\
		2 & Down Sampling Block &                $256\times 16$ & $128 \times 64$ \\
		3-7 & $5 \times$SFRB(dilation rate 1) & $128\times 64$ & $128\times 64$  \\
		8 & Down Sampling Block &                $128\times 64$ & $64\times 128$  \\ 
		9 & SFRB (dilation rate 1)&              $64\times 128$ & $64\times 128$  \\
		10 & SFRB (dilation rate 2)&              $64\times 128$ & $64\times 128$ \\
		11 & SFRB (dilation rate 1)&              $64\times 128$ & $64\times 128$ \\
		12 & SFRB (dilation rate 4)&              $64\times 128$ & $64\times 128$ \\
		13 & SFRB (dilation rate 1)&              $64\times 128$ & $64\times 128$ \\
		14 & SFRB (dilation rate 8)&              $64\times 128$ & $64\times 128$ \\
		15 & SFRB (dilation rate 1)&              $64\times 128$ & $64\times 128$ \\
		16 & SFRB (dilation rate 16)&              $64\times 128$ & $64\times 128$ \\
		17 & Transposed (s=2) &            $64\times 128$ & $128\times 64$ \\
		18 & SFRB (dilation rate 1) &             $128\times 64$ & $128\times 64$ \\
		19 & SFRB (dilation rate 1) &             $128\times 64$ & $128\times 64$ \\
		20 & Transposed (s=2) &            $128\times 64$ & $256\times 16$ \\
		21 & SFRB (dilation rate 1) &             $256\times 16$ & $256\times 16$ \\
		22 & SFRB (dilation rate 1) &             $256\times 16$ & $256\times 16$ \\
		23 & Transposed (s=2) &            $256\times 16$ & $512\times 2$ \\
		\hline \hline	
	\end{tabular}
\end{table}
\section{Proposed Methods}
In this section, we introduce each component of our efficient network ESFNet. 
The detailed information of ESFNet is shown in Table \ref{table_1}. Similar with most of segmentation networks, the whole framework is encoder-decoder architecture. The encoder is composed of the blocks from 1-16, it consists of three down sampling blocks, and two stages which have 5 SFRB, 8 SFRB respectively. And blocks of 17-23 form our light-weight decoder which employs transposed convolutions for upsampling and SFRB for fine-tuning. 
Our ESFNet requires low computational cost and memory consumption, and it achieves similar accuracy to the state-of-the-art without any other context modules, post-processing and additional scheme.

\subsection{Our Core Module}
\begin{figure}[!t]
	\centering
	\subfloat[]{\includegraphics[width=1.7in]{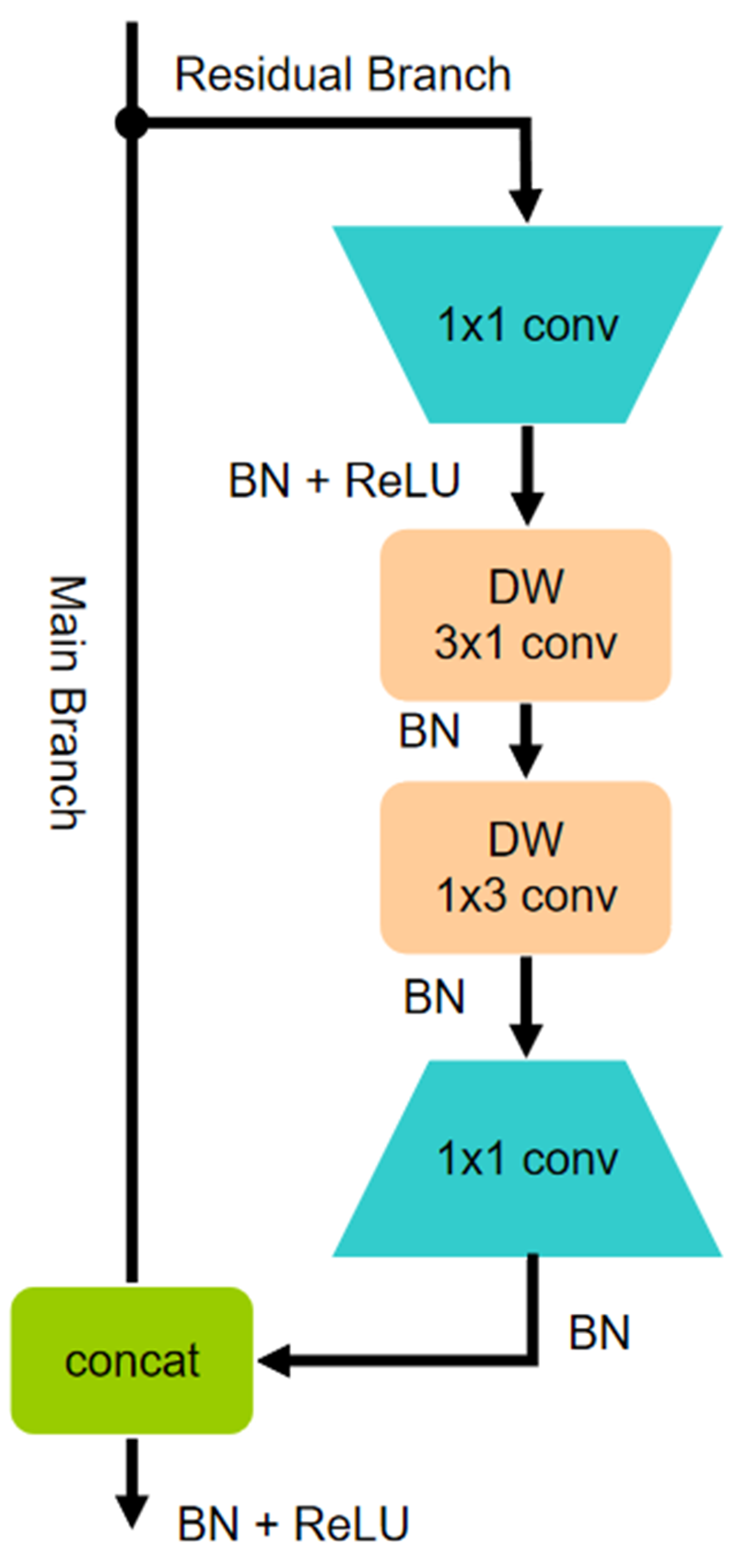}\label{fig_2a}}	
	\subfloat[]{\includegraphics[width=1.7in]{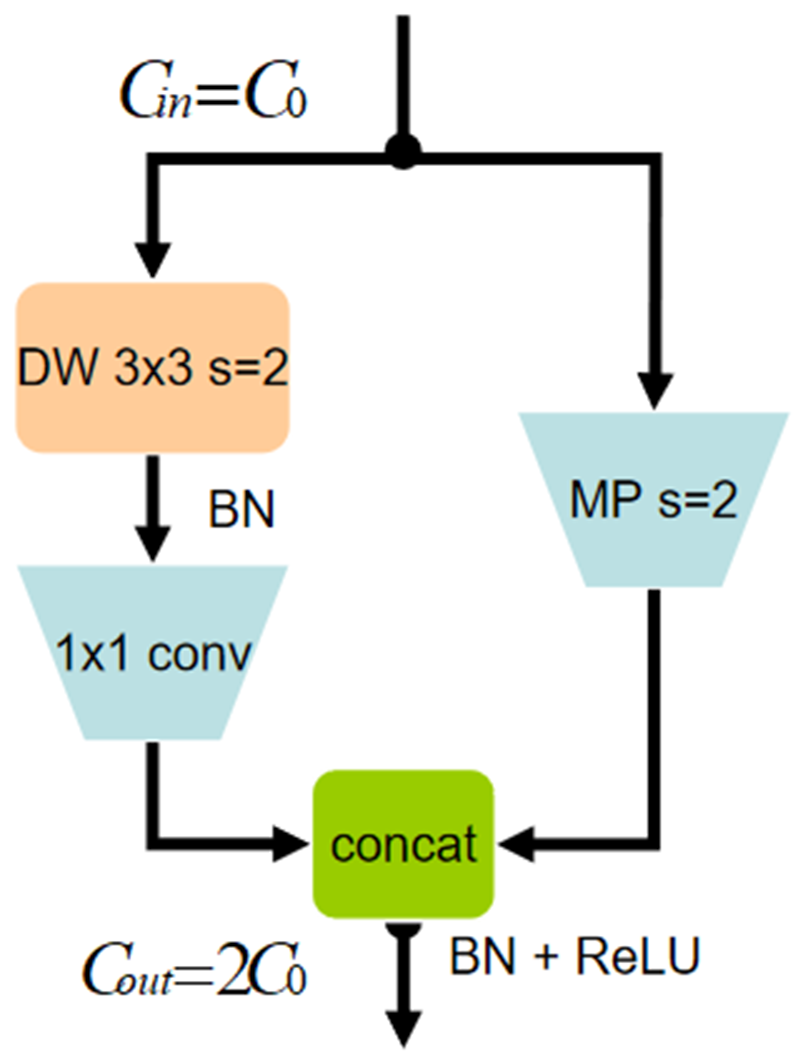}\label{fig_2b}}
	\caption{Description of our core module SFRB (a) and our down sampler (b). Where 'DW' is depth-wise convolution, 'concat' is concatenation operation, 'MP' is Map-Pooling layer and 's' is layer stride.}
	\label{fig_2}
\end{figure}

Our core module SFRB (Separable Factorized Residual Block) is shown in Fig. \ref{fig_2a}, which introduces depth-wise separable convolution into factorized residual block and incorporates with dilated convolutions to keep large receptive field with small down sampling rate. In the following part, we will explain each component of our core module SFRB in detail.
\subsubsection{Depth-wise Separable Convolutions} 
The depth-wise separable convolution is considered as key-module in recent efficient networks \cite{mobilenet, mobilenetv2, shufflenet, shufflenetv2}, it splits the full convolution operations into two independent operations, depth-wise convolution and point-wise convolution. In depth-wise convolution, the number of groups is equal to the number of feature maps, it means each kernel has single feature map in and single feature map out, and the shared weight kernels make the depth-wise convolution require less parameters than the standard. Point-wise convolution is standard convolution with kernel size of $1\times1$ , it is aimed to combine the channel-wise independent features from depth-wise convolution. Therefore, the standard convolution can be replaced by the combination of depth-wise convolution and point-wise convolution, it drastically reduces the model size and computational cost. 

For further explanation, we take $\mathbf{I}\in\mathbb{R}^{F_{in}^h\times F_{in}^w\times C_{in}}$ as input and $\mathbf{O}\in\mathbb{R}^{ F_{out}^h\times F_{out}^w\times C_{out}}$ as output, where $C_{in}$ and $C_{out}$ are the number of input channels and output channels, $F_{in}^h$, $F_{in}^w$, $F_{out}^h$ and $F_{out}^w$ are the spatial size of the input feature maps and output feature maps respectively. And weight of the convolutional layer is $\mathbf{W}\in\mathbb{R}^{C_{in}\times K_{w}\times K_{h}\times C_{out}}$, the bias is $\mathbf{b}\in\mathbb{R}^{C_{out}}$. Let $\mathbf{f_i^l}\in\mathbb{R}^{K_w\times K_h}$ denotes the $lth$ filter in $ith$ convolutional layer and $\varphi(\cdot)$ as the non-linearity. The output of $\mathbf{f_i^l}$ and the standard convolution layer can be expressed as:
\begin{equation}
\label{eqn_filter}
z(u,v)_i^l = \varphi_i[b_i^l+\sum_{c=1}^{C_{in}}\sum_{m=-\sigma}^{\sigma}\sum_{n=-\beta}^{\beta} \bar{f}_i^l \cdot \mathbf{\chi}(u+m,v+n)_{c}^T]
\end{equation}
\begin{equation}
\label{eqn_standard_conv2d}
O_i = \sum_{l=1}^{C_{out}} F_{out}^h\times F_{out}^w\times z(u,v)_i^l
\end{equation}
where $\sigma=(K_w-1)/2$, $\beta=(K_h-1)/2$, $\bar{f}_i^l$ is the 2D matrix of $lth$ filter in layer $i$, and $\mathbf{\chi}(u+m,v+n)_c$ is corresponding 2D matrix with the same size of filters and center $(u,v)$ on $cth$ feature map. From (\ref{eqn_filter}) and (\ref{eqn_standard_conv2d}), we can draw the conclusion that computational cost of standard convolution is related to spatial size of the output, kernel size and the number of input and output channels. Then the FLOPs (floating point operations) of the standard convolution can be calculated as:
\begin{equation}
\label{float_standard}
K_w\times K_h\times F_{out}^h\times F_{out}^w\times C_{in}\times C_{out}
\end{equation}
different from the standard one, the depth-wise separabel convolutions separate the full convolution into depth-wise and point-wise, it significantly reduces the computational cost and parameters needed by its shared weight scheme. Based on (\ref{eqn_filter}) and (\ref{eqn_standard_conv2d}), in depth-wise phase $C_{in}\equiv C_{out}\equiv 1$ and there are $C_{in}\times O_i$ outputs, in point-wise phase $\bar{f_i^l}$ and $\mathbf{\chi}$ are both 1D vector, so the FLOPs of the depth-wise separabel convolution is:
\begin{equation}
\label{float_depthwise}
K_w\times K_{in}\times F_{in}^h\times F_{in}^w\times C_{in} + F_{out}^h\times F_{out}^w\times C_{in}\times C_{out}
\end{equation}
for simplicity, we set $K_w\equiv K_h\equiv K$, $F_{in}^h\equiv F_{in}^w\equiv F_{out}^h\equiv F_{in}^w\equiv F$, $C_{in}\equiv C_{out}\equiv C$ and we omit the influence of bias. Therefore the standard convolution operation needs $K^2\times F^2\times C^2$ FLOPs and $K^2\times C^2$ parameters, the depth-wise separable convolution operation needs $K^2\times F^2\times C+F^2\times C^2$ FLOPs and $K^2\times C+C^2$ parameters in total. Typically, we employ kernel size of 3 in most time [17]. With that in mind, we set kernel size $K$ equal to 3 and we can calculate that there is about 9x reduction in FLOPs and the number of parameters (model size). The comparison indicates that changing standard convolutions into depth-wise separabel convolutions can make existing networks more efficient.



\subsubsection{Factorized Convolutions}
If the depth-wise convolutions decompose the standard convolutions in stage-wise, then the factorized convolutions decompose the standard convolutions in kernel-wise. As Alvarez \emph{et al}. \cite{decompsedme} has proved that, each ND kernels can be decomposed into N consecutive layers of 1D kernel. Here we only consider the 2D kernels, and it is easy to convert the situation to ND kernels. Standard 2D convolutions with $K\times K$ kernels can be converted to two 1D convolutions with $K\times 1$ kernel size and $1\times K$ kernel size. Reference \cite{learning_sep_filter} demonstrates that it is possible to relax the rank-1 constraint and essentially rewrite $f_i$ as a linear combination of 1D filters:
\begin{equation}
\label{eqn_fac_filter}
f_i = \sum_{r=1}^{r}\sigma_i^r \bar{h}_i^r (\bar{w}_i^r)^T
\end{equation}
where $\bar{h}_i^r$ and $(\bar{w}_i^r)$ are vectors of length $K$, $\sigma_i^r$ is a scalar weight, and $r$ is the rank of $f_i$. Continue with the analysis above, the $lth$ filter of 2D factorized convolution in $ith$ layer can be expressed as:
\begin{equation}
\label{eqn_fac_convolution}
z(u,v)_i^l = \varphi_i[\sum_{c=1}^{C_{in}}\sum_{-\sigma}^{\sigma}\bar{w}_i^l\varphi_i(\sum_{-\beta}^{\beta}\chi(u+m, v+n)_c^T\bar{h}_i^l)]
\end{equation}
here $m$ is in range $[-\sigma,\sigma]$, $n$ is in $[-\beta, \beta]$ and we also omit the bias. Using (\ref{eqn_standard_conv2d}) (\ref{float_standard}) (\ref{eqn_fac_filter}) (\ref{eqn_fac_convolution}), we can calculate the factorized convolution needs $2\times K\times M^2\times C^2$ FLOPs and $2\times K\times C^2$ parameters which is 3x reduction in computational cost and parameters than the standard one when $K=3$. The factorized convolutions not only shrink the models by reducing redundant parameters, but also play a role of regularizer in the whole network that can enhance generalizing capability. When incorporating with depth-wise separable convolutions, the computation and memory cost further decrease as
$2\times K\times F^2\times C + F^2\times C^2$ and $2\times K\times C + C^2$ respectively.
There are $3\times F\times F\times C$ fewer FLOPs and $3\times C$ fewer parameters than the single depth-wise separable convolution, and it is another great improvement upon convolution operation especially when the feature maps have large spatial size and amounts of channels.
\subsubsection{Dilated Convolutions and Receptive Field}
In order to improve the accuracy of segmentation in high-resolution images, the models usually need to have large RF (receptive field) \cite{duc} that can lead to rich context information for each pixel. The method used in the past to enlarge receptive field is stacking convolutional layers and down sampling layers, but too much convolution layers bring large burden of computation and memory, in addition over downsampling is harmful to dense pixel-level classification, due to the loss of unrecoverable spatial information. We evaluated 8x down sampling and 16x down sampling with the same architecture on WHU Building Dataset, the prediction of 16x down sampling network was really bad than 8x one, it proofed the spatial information is equally important with context information in dense classification tasks. Contrast to the previous methods, dilated convolution can enlarge receptive field without additional parameters increase, it is implemented by inserting $R - 1$ zeros between two consecutive kernel values along each dimension \cite{duc,pspnet}, where $R$ is dilation rate. Additionally, we define $RF_0=0$ and $FeatureStride_0=1$, the receptive field of pixels in each layer can be calculated by:
\begin{equation}
\label{eqn_9}
FeatureStride_l = \prod_{i=1}^{l} stride_i
\end{equation}
\begin{equation}
\label{eqn_10}
RF_l = RF_0 + \sum_{i=1}^{l}(K_i-1)\times FeatureStride_{i-1}\times R_i
\end{equation}
where $RF_i$, $K_i$, $R_i$ and $Stride_i$ are the size of receptive field, size of kernel, dilation rate and stride of the $ith$ layer respectively. Inspired by \cite{duc} and other novel networks \cite{deeplab3, hamaguchi_dilated_conv}, the dilation rate is usually set to sequence, we deployed dilated convolutions in stage2 demonstrated in Table \ref{table_1}, and the sequence of dilation rate is 1, 2, 1, 4, 1, 8, 1, 16. Using \eqref{eqn_9} and \eqref{eqn_10}, we can calculate the receptive field of our ESFNet is 599, and if we remove the dilated convolutions in stage2, the receptive field is only 183 that is not enough to cover the whole image. So the dilated convolution is helpful to enlarge the receptive field and enhance the performance of the networks, but it makes networks need to keep too much high-resolution feature maps in the intermediary layers, which usually leads to high memory cost \cite{denseunet}.
\subsubsection{Bottleneck or Non-bottleneck}
ResNet is designed to reduce degradation problem present in architectures that stacking large amounts of layers \cite{resnet}. There are two implementations of residual block, the bottleneck structure and non-bottleneck structure. The previous works \cite{wide_resnet} reported that non-bottleneck layers gain more accuracy from increased depth than the bottleneck one, and it indicates the bottleneck design still struggles with degradation problem. There is only around 1\% accuracy gap between the same network with non-bottleneck structure and bottleneck structure in our experiments, but the bottleneck one requires fewer parameters and lower computational complexity as shown in Table \ref{table_2}. This can be seen it does not suffer from the degradation problem, and the purpose of our work is to design efficient network with as low memory footprint and computational cost as possible, so we choose bottleneck structure as our residual block backbone. The detailed information of different structures is concluded in Table \ref{table_2}.

\begin{table}[!t]
	\renewcommand{\arraystretch}{1.3}
	\caption{Comparison of different residual structures. 'BT' is bottleneck structure, 'FAC' is factorized convolution, 'DW' is depth-wise convolutions, 'NON-BT' is non-bottleneck. The input and output are both size of $64\times 64\times 128$.}
	\label{table_2}
	\centering
	\begin{tabular}{c c c}
		\hline \hline
		\bfseries Structures & \bfseries FLOPs (M) & \bfseries Parameters (K)    \\
		\hline
		Bt & 72.09 & 17.1792 \\
		Bt+Fac & 59.64 & 14.784 \\
		Bt+Dw & 35.52 & 8.864 \\
		Bt+Fac+Dw & 35.26 & 8.832 \\
		Non-bt & 1209.01 & 295.424 \\
		Non-bt+Fac & 807.40 & 197.120 \\
		Non-bt+Dw & 145.75 & 36.096 \\
		Non-bt+Fac+Dw & 143.65 & 35.840 \\
		\hline \hline
	\end{tabular}
\end{table}

As Fig. \ref{fig_2a} demonstrates that, in residual branch the first standard $1\times1$ convolutional layer is used to compress input feature maps into 4x less, the following two depth-wise factorized convolution replace the standard $3\times3$ convolution in original work \cite{resnet} and the compressed feature maps are expanded by the last $1\times1$ convolutional layer. Note that we use batch normalization \cite{bn} and ReLU \cite{relu} non-linearity function after each convolutional layers, but no ReLU after depth-wise convolutions \cite{mobilenetv2, xception} and the $1\times1$ convolutions which are used for expanding. 
\subsection{Architecture Design}
In this subsection, we will introduce the designation of downsampling block and decoder. 
\subsubsection{Down Sampling Block}
Down sampling layers benefit for enlarging the receptive field so that each pixel can get rich context information, and it also helps to save parameters that can reduce size of model. But too much down sampling layers increase difficulty in recovering spatial information, especially harmful to dense classification. Following with the principles of ENet \cite{enet}, our networks have three down sampling layers in total, the first two is performed at very first consecutively and another down sampling layer is after stage1, we adopted the initial block in ENet as our initial block and introduced depth-wise convolution in the original as the other two down sampler which is much lighter (shown in Fig. \ref{fig_2b}). Because the output of such down samplers is concatenation of two branches, we tried to combine and refine the channel-wise independent features. Due to the great cost of using point-wise convolutions \cite{mobilenet}, we added channel shuffle at the end of down sampling blocks. But the accuracy decreases instead, so we did not use any additional feature fusion scheme after down sampling block in our final networks.
\subsubsection{Decoder Designation}
In encoder-decoder architectures, the encoder is used to extract high-level semantic features and the decoder is used to recover resolution of output from encoder to the original. The existing works \cite{segnet, unet} usually have heavy-weight decoder. Inspired by the view of light-weight decoder \cite{enet}, we tried to shorter the decoder, and we evaluated different transposed convolutions with 2x, 4x and 8x upsampling rate, but we found that the bigger upsampling rate is, the worse prediction will be got. For the shortest decoder, we removed the decoder and used the bilinear interpolation to upsample the feature maps by a factor of 8, but there was about 3 points accuracy loss. So we used transposed convolution to perform upsample with factor of 2 and used SFRB to refine the score maps in our decoder, which can be seen a better trade-off.

\section{Experiment}
In this section, we first introduce the dataset and training environment used in our experiments, and then we design sets of ablation studies to demonstrate our model is high accuracy and efficiency.
\subsection{Generl Setup}
\subsubsection{WHU Building Dataset}
We evaluated our methods on aerial imagery dataset from WHU Building Dataset which is a recent challenging dataset created by \cite{whu_dataset}. The aerial dataset consists of more than 22,000 independent buildings extracted from aerial images with 0.0075 $m$ spatial resolution and 450 $km^2$ covering in Christchurch, New Zealand. The most parts of aerial images are downsampled to 0.3 $m$ ground resolution and cropped  into 8,189 non-overlapping tiles with $512\times 512$ , theses tiles make up the whole dataset. Then it is split into three parts, 4,736 tiles for training, 1,036 tiles for validation and 2,416 tiles for testing. We train our network on the train set, valid it at the end of each epoch and test its performance on test set. The metric to measure accuracy of segmentation prediction is IoU (Intersection-over-Union) which is also called Jaccard Index and extensively adopted in segmentation tasks:
\begin{equation}
\label{eqn_11}
\frac{TP}{TP+TN+FP}
\end{equation}
Where $TP$, $FP$ and $FN$ are the number of true positives, false positives and false negatives pixels respectively.
\subsubsection{Training Configuration}
All the experiments are implemented by Pytorch1.0.1.post2 with CUDA9.0 and CuDNN7. The models are trained on single Tesla V100 in 300 epochs, using Adam optimizer with weight decay of 0.0002 and momentum of 0.9. We set the initial learning rate to 0.0005 and use poly learning rate policy as many previous works, where the learning rate is multiplied by $(1-\frac{iter}{iter_{max}})^{power}$ with power of 0.9. We set the batch size to 16 to fit our GPU memory. For fair comparisons, we also adopted the weighted loss scheme with cross entropy loss to counterwork the problem of unbalanced data, which is defined as $W_{class} = \frac{1}{\log (P_{class}+c)}$ and we set $c$ to 1.12. The data augmentation strategies we used only include random horizontal flip and random vertical flip. The whole training phase is really fast, just needs 4 hours and very low memory consumption.

\begin{table}[!t]
	\renewcommand{\arraystretch}{1.3}
	\caption{The comparison results of ablation studies. For comparing subtle difference, we adopted different precision representation which is the same in Table \ref{table_4}.}
	\label{table_3}
	\centering
	\begin{tabular}{c c c c}
		\hline\hline
		\bfseries Networks & \bfseries FLOPs (G) & \bfseries Parameters (M) &\bfseries IoU  \\
		\hline
		ESFNet-base (ours) & 2.514 & 0.1775 & 85.34 \\
		ESF-Bt & 3.075 & 0.26 & 85.63 \\
		ESF-Bt-Fac & 2.884 & 0.24 & 85.73 \\
		ESF-Bt-Dw & 2.521 & 0.1779 & 85.75 \\
		ESF-NonBt & 20.700 & 2.98 & 86.72 \\
		ESF-NonBt-Fac & 14.486 & 2.02 & 86.93 \\
		ESF-NonBt-Dw & 4.330 & 0.4495 & 86.93 \\
		ESF-NonBt-Fac-Dw & 4.275 & 0.4465 & 85.54 \\
		ESF-ENet-down & 2.744 & 0.22 & 85.56 \\
		ESF-shuffle-down & 2.513 & 0.18 & 83.30 \\
		ESF-trans2x4x & 2.112 & 0.17 & 50.58 \\
		ESF-trans8x & 1.131 & 0.10 & 13.20 \\
		ESF-interp8x & 0.527 & 0.09 & 82.12 \\
		ESF-mini & 2.373 & 0.14 & 84.57 \\
		ESF-mini-ex & 2.299 &  0.14 & 84.29 \\
		\hline\hline
	\end{tabular}
\end{table}

\subsection{Ablation Studies}
In this subsection, we conduct sets of experiments to demonstrate high performance of our methods as shown in Table \ref{table_3}. All the following results are based on test dataset.
\subsubsection{Core Modules}
The key-modules of SFRB are depth-wise separable convolutions and factorized convolutions, so we designed another two variants of our network to further explain the efficiency of our modules. The first one is “non-depth-wise” variant which replaces the standard $3\times 3$ convolution into $3\times 1$ and $1\times 3$ convolutions in bottleneck structure. The other one is “non-factorized” variant implemented by replacing the standard $3\times 3$ convolution into a depth-wise one without factorized kernels. For fair, we used the same backbone and training configuration to train these two variants, and these two networks are named as ESF-Bt-Fac and ESF-Bt-Dw respectively.

As 2-8 lines demonstrates that, our ESFNet-base achieves no more than 1\% accuracy loss compared with ESF-Bt-Fac and ESF-Bt-Dw. But the ESF-Bt-Fac has 11\% more computational cost and 30\% more parameters than ESFNet-base. The ESF-Bt-Dw is slightly more than ESFNet-base in FLOPs and the number of parameters, because the model is already small that the difference of $3\times M\times M\times C$ and $3\times C$ cannot be vast gaps. The comparison results proof the depth-wise convolutions have more powerful performance than the factorized convolutions in improving network efficiency and further combining two approaches will get much better result. The similar accuracy shows that the more parameters do not mean more accuracy, and there are much redundancies of parameters and computation in deep neural networks which need to solve urgently. 
\subsubsection{Down Sampling Block}
Following with ERFNet, we adopted the initial block as our first down sampling block, and we extended the initial block to two variants. The first variant, which is used in our final model, is implemented by replacing the standard  $3\times 3$ convolution into depth-wise one. Another one is designed for combining the channel-wise independent features, as described in Section III-B we added channel shuffle operation after the concatenation of main branch and pooling branch.

In order to investigate our efficient down sampling block, we compared ours with another two variants of ESFNet shown in 9-10 lines. The first variant named as ESF-ENet-down uses initial block in all down sampling layers likes ERFNet. The second one named as ESF-shuffle-down is implemented by adding channel shuffle operation after the concatenation at the end of down sampler. Our ESFNet-base achieves almost equal accuracy to ESF-ENet-down, but needs 18\% less parameters and 7\% less FLOPs. The basic idea of ESF-shuffle-down is to combine the channel-wise independent features, but it has 2 points accuracy drop caused by channel shuffle operation among unequal groups. From the above, the down sampler in our final networks is high accuracy and efficiency.
\subsubsection{Light-weight Decoder}
Encoder-decoder architecture is one of the most popular architectures in semantic segmentation tasks. The symmetric structure means that decoder is an exact mirror of the encoder which is deep and wide. In contrast, \cite{enet} provides a view of light-weight decoder that the decoder is only used for recovering resolution and fine-tuning. So we have an exploratory work on light-weight decoder for further studies, we performed comparisons on four variants of ESFNet with lighter and lighter decoder, they are ESFNet, ESFNet-trans2x4x, ESFNet-trans8x and ESFNet-interp8x where 2x, 4x and 8x mean the upsampling rate of transposed convolutions and interp8x means that using bilinear interpolate with factor of 8 instead of decoder.

The comparison results are reported in 11-13 lines. Though using transposed convolution with high upsampling rate like 4x and 8x brings much lighter decoder than the ESFNet-base, the accuracy drastically falls to 50.58\% and 13.20\%. Transposed convolutions implement upsampling by inserting blanks between consecutive pixels in original feature maps and then performing standard convolution operation on the upsampled feature maps which is similar to dilated convolutions. So the bad performance of transposed convolution with high upsampling rate is caused by inserting too much blanks that destroys the high-level feature representations. The bilinear interpolate is a straight-forward way to recover original resolution, the ESFNet-interp8x requires $1/5$ and $1/2$ parameters of ESFNet-base with 3 points drop in accuracy. The result of ESFNet-interp8x fits the view that the decoder is just for recovering and fine-tuning. The comparison results demonstrate the decoder in most symmetric structures \cite{segnet, deconv, unet} is "bottleneck" that effects the efficiency of networks, and it is necessary to have light-weight decoder. For preserving better balance, we employed transposed convolution with upsampling rate of 2 and SFRB for fine-tuning in our network.
\subsubsection{Mini-versions}
According to the characteristics of ResNet, we extended our ESFNet-base to mini-versions. Veit \emph{et al}. \cite{rethink_res} proposed that dropping a single residual block only has a little influence on the accuracy. Greff \emph{et al}. \cite{highway} said each residual in the same stage does not learn a new representation but learn an unrolled iterative estimation, in another word the following residual blocks is used to refine the coarse estimation from the first block in each stage. Under the premise of enough receptive fields, we dropped four SFRB in stage2 whose dilation rate is equal to one, and consider this network as ESF-mini, in this case the receptive field is 535. Furthermore, we dropped another two SFRB in stage1 and named it as ESF-mini-ex, and the receptive of ESF-mini-ex is 519 that is still enough to cover the whole image. On one hand there is too much redundancy information in early stage, on the other hand it makes the stage2 relatively close to the supervisions and lets it learn a better representation \cite{exfuse}. The ESF-mini and ESF-mini-ex have both 22\% less parameters and 8\%, 12\% fewer FLOPs compared to ESFNet-base and the accuracy loss only at around 1\%. It is worth to note that ESF-mini-ex have significant improvement on inference speed, we will discuss it in the next part.

\begin{table}[!t]
	
	\renewcommand{\arraystretch}{1.3}
	\caption{The comparison results of our architecture to the state-of-the-art networks.}
	\label{table_4}
	\centering
	\begin{tabular}{c c c c c}
		\hline\hline
		\bfseries Networks & \bfseries FLOPS (G) & \bfseries Parameters (M) &\bfseries IoU &\bfseries FPS \\
		\hline
		CU-Net & - & - & 87.1 &- \\
		FCN & - & $>$ 130 & 85.4 & - \\
		SegNet & 160.323 & 29.44 & - & 51.31 \\
		UNet & 247.532 & 26.78 & 86.8 & 35.72 \\
		SiU-Net & $>$ 495 & 26.78 & 88.4 & - \\
		ENet & 2.215 & 0.36 & 86.03 & 48.08 \\
		ERFNet & 14.674 & 2.06 & 87.03 & 88.24 \\
		EDANet & 4.410 & 0.68 & 84.05 & 88.83 \\
		Ours-base & 2.513 & 0.18 & 85.34 & 100.29 \\
		Ours-mini & 2.372 & 0.14 & 84.57 & 119.89 \\
		Ours-mini-ex & 2.299 & 0.14 & 84.29 & 142.98 \\
		\hline\hline
	\end{tabular}
\end{table}

\begin{figure*}[!t]
	\centering
	\subfloat[]{\includegraphics[width=0.9in]{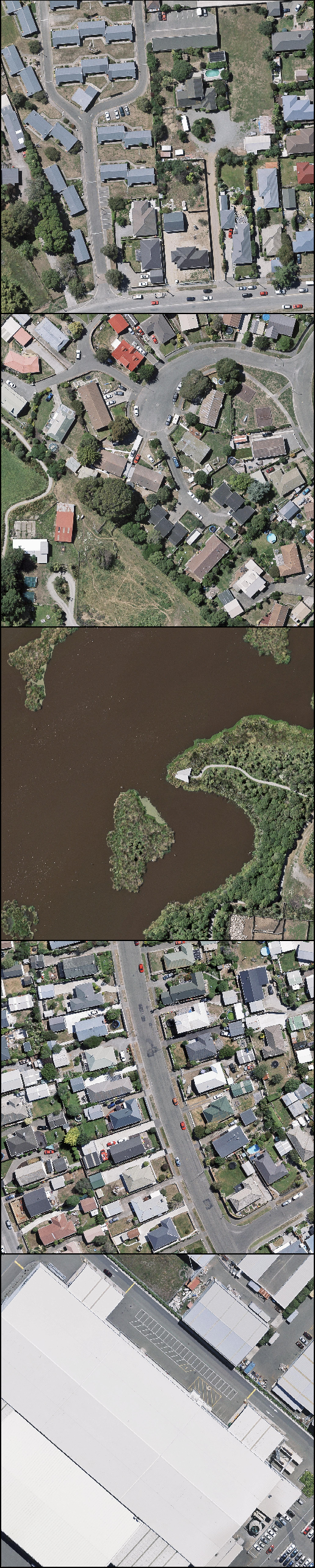}
		\label{fig_3a}}
	\hspace{-.08in}
	\subfloat[]{\includegraphics[width=0.9in]{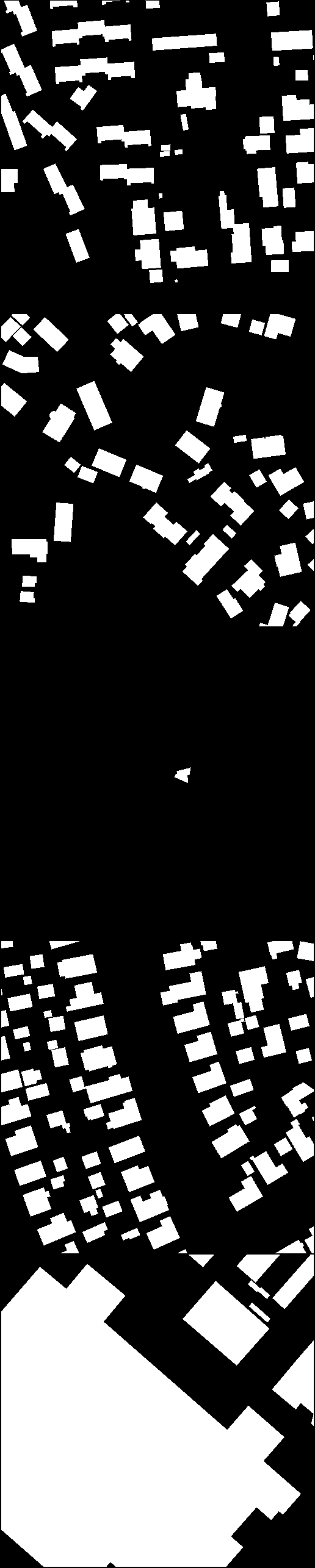}
		\label{fig_3b}}
	\hspace{-.08in}
	\subfloat[]{\includegraphics[width=0.9in]{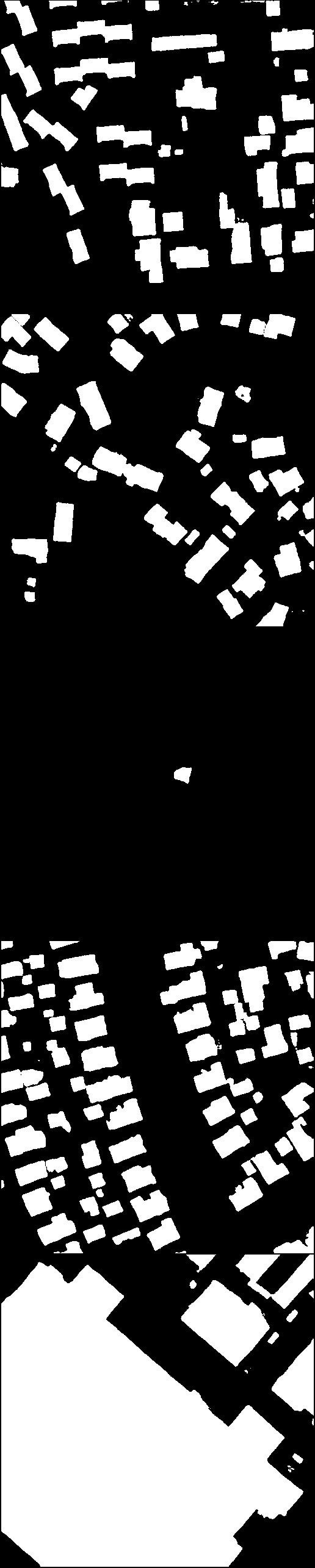}
		\label{fig_3c}}
	\hspace{-.08in}
	\subfloat[]{\includegraphics[width=0.9in]{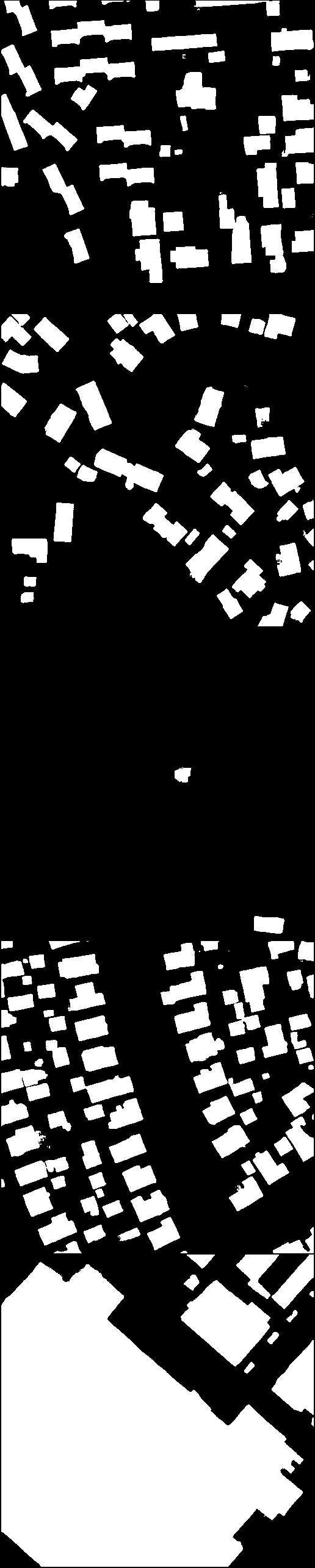}
		\label{fig_3d}}
	\hspace{-.08in}
	\subfloat[]{\includegraphics[width=0.9in]{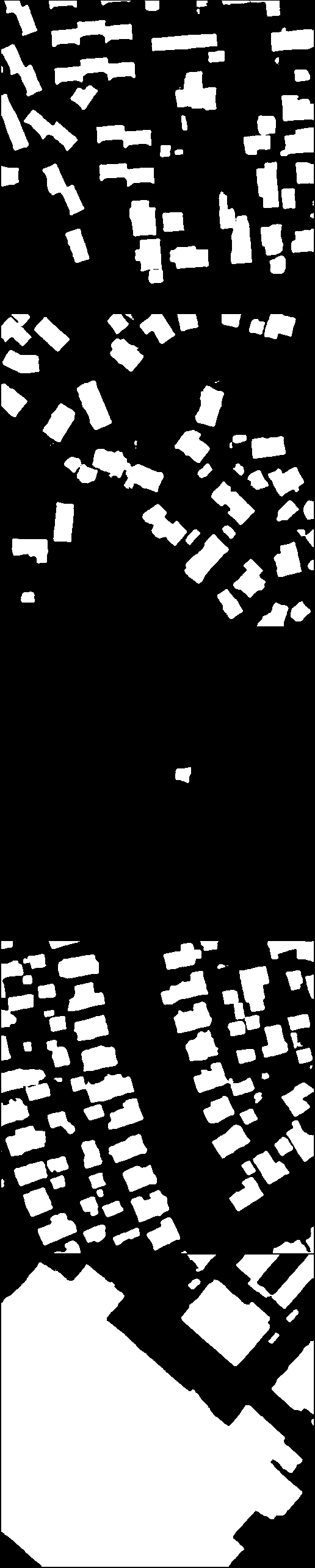}
		\label{fig_3f}}
	\hspace{-.08in}
	\subfloat[]{\includegraphics[width=0.9in]{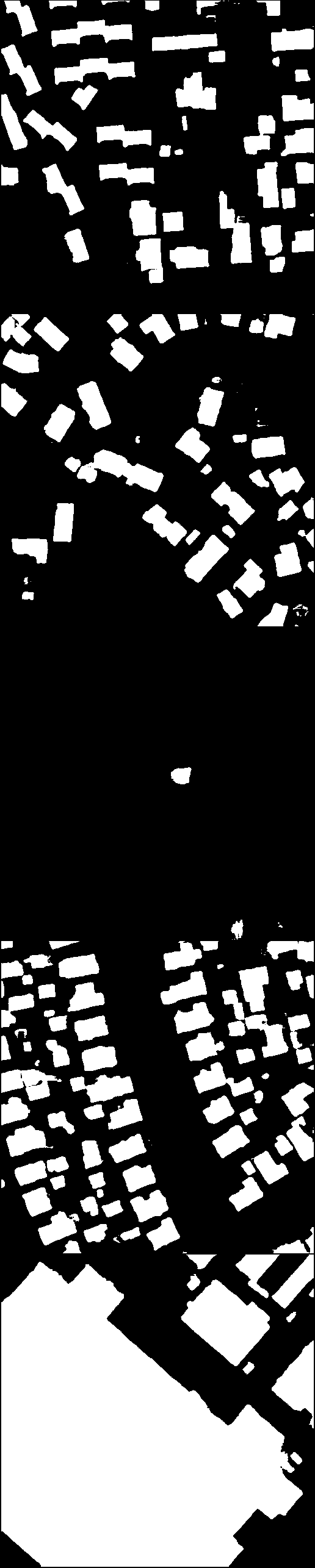}
		\label{fig_3e}}
	\hspace{-.08in}
	\caption{Prediction results of ENet (c), ERFNet (d), our ESFNet (e) and ESFNet-mini-ex (f). (a) Aerial images. (b) Ground Truth.}
	\label{fig_3}
\end{figure*}
\section{Evaluation}
As shown in Table \ref{table_4}, we compared our ESFNet-base and its mini-versions with other state-of-the-art networks (i.e. SiU-Net, CU-Net, UNet and FCN which come from the original paper \cite{whu_dataset}) and novel real-time architectures (i.e. ENet, ERFNet and EDANet which are in similar structure and easy to deploy). All the comparison results are based on test set with the same training environment and configuration. The accuracy of networks is measured by IoU, and we used FPS (Frames Per Second) to measure the inference speed, and “-” means that this value is not given. The architectures of CU-Net and FCN are not given clearly in the original paper \cite{whu_dataset}, so we do not display their FLOPs, Params and FPS. But we know CU-Net has similar structure with UNet and FCN-8s \cite{fcn} already has over 130 M parameters, both of them are heavy-weight networks. Though SegNet has much lighter decoder than other encoder-decoder architectures, it still has so many parameters of 29.44 M that we cannot train it on a single GPU with batch size of 16 and we do not display its IoU. 

The results show that our ESFNet-base can achieve similar accuracy to the state-of-the-art and obtain much better trade-off between accuracy and efficiency than previous real-time networks. Our ESFNet-base can run 100.29 FPS, it achieves 85.34\% IoU on test dataset with 2.513 G FLOPs and 0.18 M parameters. We extended our ESFNet-base to ESF-mini-ex which has only 1\% accuracy drop but with lower FLOPs and less parameters than the base, the ESFNet-mini-ex highly improves the inference speed to 142.98 FPS.

\vspace{-0.5em}
Most of previous networks for segmentation tasks are encoder-decoder architecture, by making networks deeper and wider or using additional scheme to achieve high accuracy, such as SegNet, SiU-Net, CU-Net and UNet, but these networks need high computation and memory resources. The top-accuracy method SiU-Net in Table \ref{table_4}, which uses the multi-pipline input with shared weight UNet, achieves 88.4\% IoU but requires 26.78 M parameters and more than 495 G FLOPs. Compared to the top-accuracy network, our ESFNet-base only has accuracy drop of 3 points with 148x less parameters 196x less FLOPs. Our ESFNet-base achieves similar accuracy to the FCN, but FCN has over 130 M parameters which is a huge number compared to 0.18 M. It demonstrates that there is large amount of redundancies in previous high-accuracy networks, and our networks can utilize parameters much more efficiently to learn the similar representation to the top-accuracy one. In summary, previous high-accuracy networks are not comparable to our networks and other real-time networks in efficiency, and our networks can achieve better trade-off between accuracy and speed.

We compared our ESFNet to recent real-time networks, which both utilize residual learning. The ENet and ERFNet both benefits from ResNet architecture, ERFNet achieves much higher IoU 87.03\% than ENet 86.03\% but it can run 1.8x faster than ENet with higher FLOPs and lager model size in our experiment. It proofs the performance of factorized convolutions both in accuracy and efficiency. EDANet has lower FLOPs and less parameters than ERFNet, it achieves 84.05\% IoU but with similar inference speed to ERFNet, the reason is that Dense-Blocks need much memory resources to save high resolution feature maps in intermediate layers which leads too much cost of memory access, and the lower accuracy is caused by bilinear interpolate upsampling method as analyzed in Section IV-B. Our ESFNet-base achieves 85.32\% IoU on test set and it can run 12 more frames per second than ERFNet. Our ESFNet-mini-ex highly improves the inference speed to 142.98 FPS while preserving acceptable accuracy of 84.29\% IoU. Therefore, our core module SFRB and our network designation scheme significantly enhances the performance of networks for semantic segmentation tasks, both in accuracy and efficiency.

The comparison of prediction results is shown in Fig. \ref{fig_3}. To test the performance of our ESFNet, we specifically selected the aerial images in which the buildings are small, large or in fragmentation. Our network performed well on small buildings (the third row) and large buildings (the fifth row), but there is slight loss of precision in extremely tiny buildings (bottom left of the second row) and building boundaries. The comparison result demonstrates the proposed ESFNet can get similar predictions compared to the state-of-the-art, but our ESFNet implements the better trade-off between accuracy and efficiency.

\section{Conclusions}
In this paper, we proposed novel efficient and accurate architecture ESFNet for building extraction task. Our ESFNet-base achieves 85.34\% IoU and 100.29 FPS on WHU Building Dataset, and our ESFNet-mini-ex achieves 84.29\% IoU and 142.98 FPS. We designed a highly efficient module SFRB as our core module which can be deployed in most of existing architectures. Through comprehensive ablation studies and sets of comparisons with state-of-the-art, we analyzed the efficient and accurate network designation scheme for semantic segmentation networks. In summary, our ESFNet can provide the better trade-off between accuracy and efficiency that makes remote sensing researches much more efficient.

During the experiments, we observed some unexpected results which are different from the previous works. We found the memory access cost is another factor that can influence the inference speed. In the future work, we will interleave the memory access cost with the current metrics to analyze the efficiency of deep neural networks. For deploying in practical applications, we will further evaluate our networks on other embedded devices.

\ifCLASSOPTIONcaptionsoff
  \newpage
\fi



\bibliographystyle{IEEEtran}
\end{document}